# Learning Polytrees

**Sanjoy Dasgupta**
Department of Computer Science
University of California, Berkeley

## Abstract

We consider the task of learning the maximum-likelihood polytree from data. Our first result is a performance guarantee establishing that the optimal branching (or Chow-Liu tree), which can be computed very easily, constitutes a good approximation to the best polytree. We then show that it is not possible to do very much better, since the learning problem is NP-hard even to approximately solve within some constant factor.

## 1 Introduction

The huge popularity of probabilistic nets has created a pressing need for good algorithms that learn these structures from data. Currently the theoretical terrain of this problem has two major landmarks. First, it was shown by Edmonds (1967) that there is a very simple greedy algorithm for learning the maximum-likelihood *branching* (undirected tree, or equivalently, directed tree in which each node has at most one parent). This algorithm was further simplified by Chow and Liu (1968), and as a result branchings are often called *Chow-Liu trees* in the literature. Second, Chickering (1996) showed that learning the structure of a general directed probabilistic net is NP-hard, even if each node is constrained to have at most two parents. His learning model uses Bayesian scoring, but the proof can be modified to accommodate a maximum-likelihood framework.

With these two endposts in mind, it is natural to ask whether there is some subclass of probabilistic nets which is more general than branchings, and for which a provably good structure-learning algorithm can be found. Even if this learning problem is NP-hard, it may well be the case that a good approximation algorithm exists – that is, an algorithm whose answer is guaranteed to be close to optimal, in terms of log-likelihood or some other scoring function. We emphasize that from a theoretical standpoint an NP-hardness result closes one door but opens many others. Over the last two decades, provably good approximation algorithms have been developed for a wide host of NP-hard optimization problems, including for instance the Euclidean travelling salesman problem. Are there approximation algorithms for learning structure?

We take the first step in this direction by considering the task of learning polytrees from data. A *polytree* is a directed acyclic graph with the property that ignoring the directions on edges yields a graph with no undirected cycles. Polytrees have more expressive power than branchings, and have turned out to be a very important class of directed probabilistic nets, largely because they permit fast exact inference (Pearl, 1988). In the setup we will consider, we are given i.i.d. samples from an unknown distribution $D$ over $(X_1, \ldots, X_n)$ and we wish to find some polytree with $n$ nodes, one per $X_i$, which models the data well. There are no assumptions about $D$.

The question "how many samples are needed so that a good fit to the empirical distribution reflects a good fit to $D$?" has been considered elsewhere (Dasgupta, 1997; Höffgen, 1993), and so we will not dwell upon sample complexity issues here.

How do we evaluate the quality of a solution? A natural choice, and one which is very convenient for these factored distributions, is to rate each candidate solution $M$ by its log-likelihood. If $M$ assigns parents $\Pi_i$ to variable $X_i$, then the maximum-likelihood choice is simply to set the conditional probabilities $P(X_i = x_i \mid \Pi_i = \pi_i)$ to their empirically observed values. The negative log-likelihood of $M$ is then

$$-ll(M) = \sum_{i=1}^{n} H(X_i|\Pi_i),$$

where $H(X_i|\Pi_i)$ is the conditional entropy of $X_i$ given $\Pi_i$ (with respect to the empirical distribution), that is to say the randomness left in $X_i$ once the values of its parents $\Pi_i$ are known. A discussion of entropy and related quantities can be found in the book of Cover and Thomas (1991). The core combinatorial problem can now be defined as

*PT: Select a (possibly empty) set of parents $\Pi_i$ for each node $X_i$, so that the resulting structure is a polytree and so as to minimize the quantity $\sum_{i=1}^{n} H(X_i|\Pi_i)$,* or in deference to sample complexity considerations,

*PT(k): just like PT, except that each node is allowed at most $k$ parents (we shall call these $k$-polytrees).*

It is not at all clear that PT or PT($k$) is any easier than learning the structure of a general probabilistic net. However, we are able to obtain two results for this problem. First, the optimal branching (which can be found very quickly) constitutes a provably good approximation to the best polytree. Second, it is not possible to do very much better, because there is some constant $c > 1$ such that it is NP-hard to find any 2-polytree whose cost (negative log-likelihood) is within a multiplicative factor $c$ of optimal. That is to say, if on input $I$ the optimal solution to PT(2) has cost $OPT(I)$, then it is NP-hard to find any 2-polytree whose cost is $\leq c \cdot OPT(I)$. To the best of our knowledge, this is the first result that treats hardness of approximation in the context of learning structure.

Although our positive result may seem surprising because the class of branchings is less expressive than the class of general polytrees, it is rather good news. Branchings have the tremendous advantage of being easy to learn, with low sample complexity requirements – since each node has at most one parent, only pairwise correlations need to be estimated – and this has made them the centerpiece of some very interesting work in pattern recognition. Recently, for instance, Friedman, Geiger, and Goldszmidt (1997) have used branchings to construct classifiers, and Meilă, Jordan, and Morris (1998) have modelled distributions by mixtures of branchings.

There has been a lot of work on reconstructing polytrees given data which actually comes from a polytree distribution – see, for instance, the papers of Geiger, Paz and Pearl (1990) and Acid, de Campos, González, Molina and Pérez de la Blanca (1991). In our framework we are trying to approximate an *arbitrary* distribution using polytrees. Although there are various local search techniques, such as EM or gradient descent, which can be used for this problem, no performance guarantees are known for them. Such guarantees are very helpful, because they provide a scale along which different algorithms can be meaningfully compared, and because they give the user some indication of how much confidence he can have in the solution.

The reader who seeks a better intuitive understanding of the polytree learning problem might find it useful to study the techniques used in the performance guarantee and the hardness result. Such proofs inevitably bring into sharp focus exactly those aspects of a problem which make it difficult, and thereby provide valuable insight into the hurdles that must be overcome by good algorithms.

## 2  An approximation algorithm

We will show that the optimal branching is not too far behind the optimal polytree. This instantly gives us an $O(n^2)$ approximation algorithm for learning polytrees.

Let us start by reviewing our expression for the *cost* (negative log-likelihood) of a candidate solution:

$$-ll(M) = \sum_{i=1}^{n} H(X_i|\Pi_i). \qquad (\dagger)$$

This simple additive scoring function has a very pleasing intuitive explanation. The distribution to be modelled (call it $D$) has a certain inherent randomness content $H(D)$. There is no structure whose score is less than this. The most naive structure, which has $n$ nodes and no edges, will have score $\sum_i H(X_i)$, where $H(X_i)$ is the entropy (randomness) of the individual variable $X_i$. There is likewise no structure which can possibly do worse than this. We can think of this as a starting point. The goal then is to carefully add edges to decrease the entropy-cost as much as possible, while maintaining a polytree structure. The following examples illustrate why the optimization problem might not be easy.

**Example 1.** Suppose $X_1$ by itself looks like a random coin flip but its value is determined completely by $X_2$ and $X_3$. That is, $H(X_1) = 1$ but $H(X_1|X_2, X_3) = 0$. There is then a temptation to add edges from $X_2$ and $X_3$ to $X_1$, but it might ultimately be unwise to do so because of problematic cycles that this creates. In particular, a polytree can contain at most $n - 1$ edges, which means that each node must on average have (slightly less than) one parent. ∎

**Example 2.** If $X_2, X_3$ and $X_4$ are independent random coin flips, and $X_1 = X_2 \oplus X_3 \oplus X_4$, then adding edges from $X_2, X_3$ and $X_4$ to $X_1$ will reduce the entropy of $X_1$ by one. However, any proper subset of these three parents provides no information about $X_1$; for instance, $H(X_1|X_2, X_3) = H(X_1) = 1$. ∎

The cost ($\dagger$) of a candidate solution can lie anywhere in the range $[0, O(n)]$. *The most interesting situations are those in which the optimal structure has very low cost (low entropy), since it is in these cases that there is structure to be discovered.*

We will show that the optimal branching has cost (and therefore log-likelihood) which is at most about a constant factor away from that of the optimal polytree. This is a very significant performance guarantee, given that this latter cost can be as low as one or as high as $n$.

## 2.1 A simple bound

The following definition will be crucial for the rest of the development.

**Definition.** *For a given probability distribution over variables $X_1, \ldots, X_n$, let $U = \max_i H(X_i)$ (that is, the highest entropy of an individual node) and $L = \min_i H(X_i)$.*

**Example 3.** If all the variables are boolean and each by itself seems like a random coin flip, then $U = L = 1$. Similarly, if some variables have values in the range $\{1, 2, \ldots, 100\}$ and the rest are boolean, and each variable by itself looks pretty uniform-random (within its range), then $U \approx \log_2 100 < 7$ and $L \approx 1$. The reader might want to convince himself that the cost (negative log-likelihood) of the optimal polytree (or the optimal branching, or the optimal directed probabilistic net) will be in the range $[L, nU]$. ∎

We warm up with an easy theorem which provides useful intuition about PT.

**Theorem 1.** *The cost (negative log-likelihood) of the optimal branching is at most $(1 + U/L)$ times than that of the optimal polytree.*

*Proof.* Let the optimal polytree have total cost $H^*$ (we have chosen the letter $H$ because the cost is essentially an entropy rating, and it is helpful to think of it as such). We will use this polytree to construct a branching of cost $\leq H^*(1 + U/L)$, and this will prove the theorem.

In the optimal solution, let $S$ denote the vertices with indegree more than one. Since the total number of edges in the structure is at most $n - 1$, *each node of high indegree is effectively stealing inedges away from other nodes* (cf. Example 1). In particular therefore, the polytree must have at least $|S| + 1$ sources (that is, nodes with no parents), implying that $H^* \geq |S|L$.

If we remove the edges into the vertices of $S$ (so that they become sources in the graph), we are left with a branching. Each vertex of $S$ has entropy at most $U$, and so the cost of this branching is $\leq H^* + |S|U \leq H^*(1 + U/L)$. ∎

Thus in cases where the nodes have approximately equal individual entropies, the optimal branching is at most about a factor of two worse than the optimal polytree, in terms of log-likelihood. What happens when the ratio $U/L$ is pretty large? This is especially a danger when different variables have vastly different ranges. Over the course of the next few lemmas, we will improve the dependence on $U/L$ to $O(\log U/L)$, and we will show that this is tight. We start with a very crude bound which will be useful to us later.

## 2.2 A better performance guarantee

First we establish some simple

**Notation.** Let the optimal polytree have cost $H^*$, as before. Whenever we talk about "parents" and "ancestors" henceforth, it will be with respect to this solution. We say $X$ is a parent of $Y$ if there is an edge from $X$ to $Y$; the definitions of "child" and "ancestor" follow from this. A source is a node with no inedges and a sink is a node with no outedges. Denote by $T_X$ the induced subgraph consisting of node $X$ and its ancestors. That is, $T_X$ is a (directed) subgraph of the polytree such that its only sink is $X$ and all the other nodes in it have directed paths to $X$. Let $|T_X|$ then signify the number of nodes in this subtree. For each node $Z$ with parents $\Pi$, let $\Delta(Z) = H(Z|\Pi)$, the entropy that remains in $Z$ even after its parents are known. Clearly $H^* = \sum_Z \Delta(Z)$. Extend $\Delta$ to subgraphs $T_Z$ in the natural way: $\Delta(T_Z) = \sum_{X \in T_Z} \Delta(X)$. Finally, all logarithms in this paper will be base two.

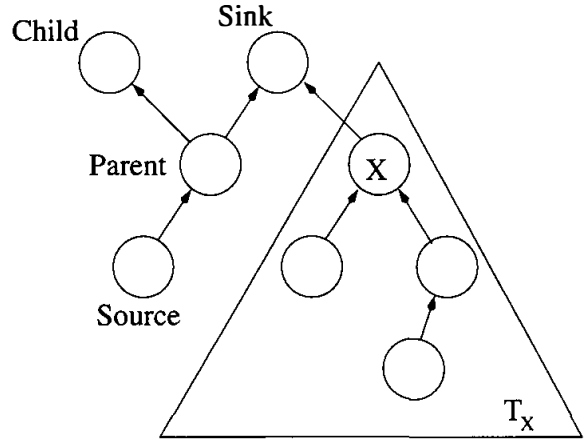

The ensuing discussion will perhaps most easily be understood if the reader thinks of $\Delta(Z)$ as some positive real-valued attribute of node $Z$ (such that the sum of these values over all nodes is $H^*$) and ignores its original definition in terms of $Z$'s parents in the optimal polytree. We start with a few examples to clarify the notation.

**Example 4.** Suppose that in the optimal polytree, node $X$ has parents $Y_1, \ldots, Y_k$. Then the subtrees $T_{Y_1}, \ldots, T_{Y_k}$ are disjoint parts of $T_X$ and therefore

$$\sum_{i=1}^{k} \Delta(T_{Y_i}) \leq \Delta(T_X) \leq H^*.$$

Decompositions of this kind will constitute the bread and butter of the proofs that follow. ∎

**Example 5.** For any node $X$, we can upper-bound its individual entropy,

$$H(X) \leq \Delta(T_X),$$

since the right-hand term is the total entropy of the subtree $T_X$ which includes $X$. Here's another way to think of it: denote by $S$ the variables (including $X$) in the subtree $T_X$. Then $T_X$ is a polytree over the variables $S$, and so $\Delta(T_X)$ is at least the inherent entropy $H(D|_S)$ of the target distribution $D$ restricted to variables $S$, which in turn is at least $H(X)$. ∎

Given the optimal polytree, we need to somehow construct a branching from it which approximates it well. Nodes with zero or one parent can remain unchanged, but for each node $Z$ with parents $Y_1, Y_2, \ldots, Y_l, l \geq 2$, we must eliminate all but one inedge. Naturally, *the parent we choose to retain should be the one which contributes the most information to $Z$*. The information lost by removing parent $Y_i$ is at most its entropy $H(Y_i)$, which in turn is upper-bounded by $\Delta(T_{Y_i})$ (see Example 5). Therefore, we will use the simple rule: *retain the parent with highest $\Delta(T_{Y_i})$*. Formally, the increase in cost occasioned by removing all parents of $Z$ except $Y_j$ can be bounded by:

$$H(Z|Y_j) - H(Z|Y_1, Y_2, \ldots, Y_l)$$
$$\leq \sum_{i \neq j} H(Y_i)$$
$$\leq \sum_{i \neq j} \Delta(T_{Y_i})$$
$$= \sum_i \Delta(T_{Y_i}) - \max_i \Delta(T_{Y_i}).$$

This motivates the following

**Definition.** *For any node $Z$ with $l \geq 2$ parents $Y_1, \ldots, Y_l$ in the optimal polytree, let the charge $C(Z)$ be $\sum_i \Delta(T_{Y_i}) - \max_i \Delta(T_{Y_i})$. For nodes with less than two parents, $C(Z) = 0$.*

So we apply our rule to each node with more than one parent, and thereby fashion a branching out of the polytree that we started with. All these edges removals are going to drive up the cost of the final structure, but by how much? Well, there is no increase for nodes with less than two parents. For other nodes $Z$, we have seen that the increase is at most $C(Z)$.

In this section, we will charge each node $Z$ exactly the amount $C(Z)$, and it will turn out that the total charges are at most $1/2 H^* \log n$. Most of the work is done in the following

**Lemma 2.** *For any node $Z$, the total charge for all nodes in subgraph $T_Z$ is at most $1/2 \Delta(T_Z) \log |T_Z|$.*

*Proof.* Let $C(T_Z) = \sum_{X \in T_Z} C(X)$ denote the total charge for nodes in subgraph $T_Z$. We will use induction on $|T_Z|$. If $|T_Z| = 1$, $Z$ is a source and trivially $C(T_Z) = 0$. So, assume the claim is true for all subtrees of size less than $|T_Z|$. If $Z$ has just one parent, say $Y$, then $C(T_Z) = C(T_Y)$, and we are done. Assume, then, that $Z$ has parents $Y_1, \ldots, Y_l, l \geq 2$, and let $\delta_i = |T_{Y_i}|/|T_Z|$. Then $\delta_1 + \cdots + \delta_l \leq 1$, and

$$\begin{aligned} C(T_Z) &= \sum_i C(T_{Y_i}) + C(Z) \\ &\leq \sum_i 1/2 \Delta(T_{Y_i}) \log \delta_i |T_Z| + \\ &\quad \sum_i \Delta(T_{Y_i}) - \max_i \Delta(T_{Y_i}) \\ &\leq 1/2 \Delta(T_Z) \log |T_Z| + \\ &\quad \sum_i \Delta(T_{Y_i})(1 + 1/2 \log \delta_i) - \max_i \Delta(T_{Y_i}) \\ &\leq 1/2 \Delta(T_Z) \log |T_Z|, \end{aligned}$$

where the second line applies the inductive hypothesis to subtrees $T_{Y_1}, \ldots, T_{Y_l}$, the third line follows from $\Delta(T_Z) \geq \sum_i \Delta(T_{Y_i})$ (because these $l$ subtrees are disjoint), and the fourth line is the result of (1) ignoring all $\delta_i$ smaller than $1/4$, (2) applying Jensen's inequality, and (3) using the fact that $\max_i a_i$ dominates any convex combination of $\{a_i\}$. ∎

Now we apply this to selected subgraphs of the optimal polytree. Our aim, again, is to bound the additional cost incurred by removing edges necessary to convert the optimal polytree into a branching. Specifically, we want to show

$$\sum_{i=1}^n C(X_i) \leq 1/2 H^* \log n = \frac{1}{2} \left( \sum_{i=1}^n \Delta(X_i) \right) \log n.$$

**Theorem 3.** *The cost of the optimal branching is at most $(1 + 1/2 \log n)$ times the cost of the optimal polytree.*

*Proof.* Suppose the optimal polytree were the union of a few disjoint subtrees $T_{Z_1}, \ldots, T_{Z_l}$ (equivalently, suppose each connected component had just one sink). Then we would be in business, because we could apply the previous lemma to each of these subtrees and then sum the results. We would get

$$\begin{aligned} \sum_{i=1}^n C(X_i) &= \sum_{i=1}^l C(T_{Z_i}) \\ &\leq \sum_{i=1}^l 1/2 \Delta(T_{Z_i}) \log |T_{Z_i}| \\ &\leq \sum_{i=1}^n 1/2 \Delta(X_i) \log n \\ &= 1/2 H^* \log n, \end{aligned}$$

where the second line uses Lemma 2 and the third line uses the fact that $|T_{Z_i}| \leq n$.

If the optimal polytree is not of this convenient form, then it must have some connected component which contains two sinks $X$ and $Y$ such that $T_X$ and $T_Y$ contain a common subgraph $T_U$.

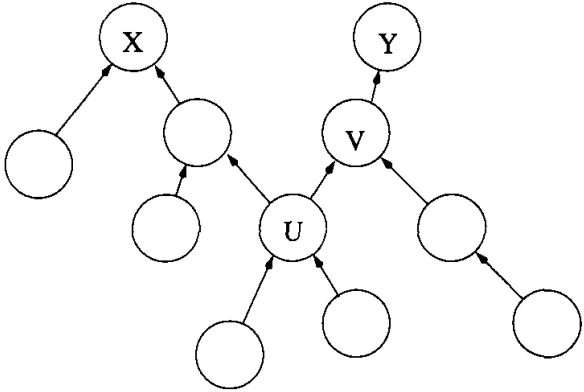

Say that $T_U$ is connected to the rest of $T_Y$ through the edge $(U, V), V \in T_Y - T_U$. Remove this edge, and instead add the edge $(X, V)$. This has the effect of moving $T_Y - T_U$ all the way up $T_X$. After this change, $X$ is no longer a sink, and for every node $Z$ in the graph, $\Delta(T_Z)$ (and thus $C(Z)$) has either stayed the same or increased, because each node has all the ancestors it had before and maybe a few more (recall that each $\Delta(W)$ is a fixed value associated with node $W$ and $\Delta(T_Z)$ is the sum of these values over $W \in T_Z$). It is also the case, and this will be useful later, that the indegree of each node remains unchanged. In this manner we alter the structure so that it is a disjoint union of single-sink components, and then apply the previous argument. ∎

**Remark.** In the optimal polytree, let $n_0$ denote the number of nodes with no parents (that is, the number of sources), and $n_{\geq 2}$ the number of nodes with at least 2 parents. By examining what happens to nodes of different indegree in the above proof, we notice that the theorem remains true if we replace $n$ by $n_0 + n_{\geq 2}$, that is, if we ignore nodes with exactly one inedge.

### 2.3 The final improvement

In the argument so far, we have bounded the entropy of a node $X$ by $H(X) \leq \Delta(T_X)$. This is a fairly tight bound in general, except when these entropy values start getting close to the maximum entropy $U$. We factor this in by using a slightly more refined upper bound.

**Definition.** For any node $Z$, let $C'(Z) = \min\{C(Z), U\}$. Then $C'(Z)$ is an upper bound on the extra cost incurred at node $Z$ due to the removal of all but one parent.

This gives us the final improvement. Once again, we start by considering a single subtree.

**Lemma 4.** *For any node $Z$, the total cost incurred by nodes in the subtree $T_Z$ is at most $C'(T_Z) \leq (5/2 + 1/2 \log U/L)\Delta(T_Z)$.*

*Proof.* Let $T \subset T_Z$ denote the polytree induced by all nodes $X \in T_Z$ such that $\Delta(T_X) > U$; that is, $T$ consists of these nodes, and any edges between them in $T_Z$. Note that $T$ must be connected, because if $X \in T$, and $Y \in T_Z$ is a child of $X$, then $Y$ must also be in $T$ (since $\Delta(T_Y) \geq \Delta(T_X)$). Let $T_j \subset T$ be those nodes of $T$ which have $j$ parents *in* $T$. And let $B$ be the border nodes right outside $T$, that is, nodes which are in $T_Z - T$ and have children in $T$.

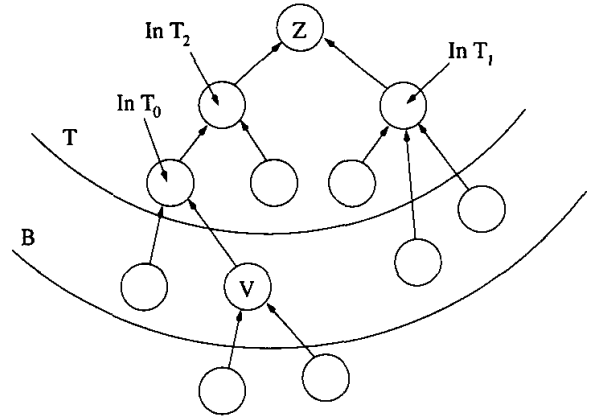

What are the charges for the various nodes in $T$?

(1) Each node $X \in T_1$ has all but one parent outside $T$, and therefore gets charged at most the sum of $\Delta(T_Y)$ over its parents $Y$ outside $T$ (and in $B$). The total charge for nodes in $T_0$ and $T_1$ is

$$\sum_{X \in T_0} C'(X) + \sum_{X \in T_1} C'(X) \leq \sum_{Y \in B} \Delta(T_Y)$$
$$\leq \Delta(T_Z)$$

where the final inequality follows by noticing that the preceding summation is over disjoint subtrees of $T_Z$ (cf. Example 4).

(2) Nodes $X \in T_2, T_3, \ldots$ have at least two parents in $T$, and so get charged exactly $U$; however, since $T$ is a tree we know that $|T_0| \geq 1 + |T_2| + |T_3| + \cdots$, and thus

$$\sum_{X \in T_i, i \geq 2} C'(X) \leq \sum_{X \in T_0} U$$
$$\leq \sum_{X \in T_0} \Delta(T_X)$$
$$\leq \Delta(T_Z)$$

where once again the disjoint subtrees property is used.

Thus nodes in $T$ have total charge at most $2\Delta(T_Z)$. It remains to assess the charges for those nodes in $T_Z$ which are not in $T$. Split these nodes into their disjoint sub-polytrees $\{T_V : V \in B\}$, and consider one such $T_V$. Since $\Delta(T_V) \leq U$ and each source in $T_V$ has entropy at least $L$, we can conclude that $T_V$ has at most $U/L$ sources and therefore at most $2U/L$ nodes of indegree $\neq 1$. The charge $C'(T_V)$ is therefore, by Lemma 2, at most $\frac{1}{2}\Delta(T_V)\log 2U/L$. We sum this over the various disjoint polytrees and find that the total charge for $T_Z - T$ is at most $\frac{1}{2}\Delta(T_Z)\log 2U/L$, whereupon $C'(T_Z) \leq (5/2 + 1/2 \log U/L)\Delta(T_Z)$, as promised. ∎

**Theorem 5.** *The cost of the optimal branching is at most $(7/2 + 1/2 \log U/L)$ times the cost of the optimal polytree.*

*Proof.* It can be verified as in Theorem 3 that for charging purposes we may assume the optimal solution is a disjoint union of single-sink subgraphs. Apply the previous lemma to each of these subtrees in turn, and sum the results. ∎

This immediately gives us a reasonable approximation algorithm for PT.

The bound on the ratio between the best branching and the best polytree has to depend upon $\log U/L$. To see this, consider a polytree which is structured as a complete binary tree (with $n/2 + 1$ leaves) and with edges pointing towards the single sink. All nodes are binary-valued, each internal node is the exclusive-or of its two parents, and each source has some small entropy $\epsilon = \theta(1/n)$. The nodes on the next level up each have approximately double this entropy, and so on, so that the sink has entropy about $\theta(1)$. The optimal polytree has cost $(n/2 + 1)\epsilon = \theta(1)$ whereas the best branching loses $\theta(\epsilon n)$ entropy on each level and therefore has cost about $\theta(\epsilon n \log n) = \theta(\log n) = \theta(\log U/L)$.

## 3  A hardness result

Needless to say, the decision version of PT is NP-complete. However, the news is considerably worse than this – we will prove a hardness barrier which rules out the possibility of a polynomial time approximation scheme. Edmonds' algorithm solves PT(1) optimally; it will now turn out that if nodes are allowed a second parent then the problem becomes hard to approximately solve.

**Theorem 6.** *There is some constant $c > 1$ for which the problem of finding a 2-polytree whose cost is within $c$ times optimal is an NP-hard optimization problem.*

*Proof.* We shall reduce from the following canonical problem ($\epsilon > 0$ is some fixed constant).

MAX3SAT(3)
*Input:* A Boolean formula $\phi$ in conjunctive normal form, in which each clause contains at most three literals and each variable appears exactly three times. It is guaranteed that either there is an assignment which satisfies all the clauses (that is, $\phi$ is satisfiable) or no assignment satisfies more than $(1 - \epsilon)$ fraction of the clauses.
*Question:* Is $\phi$ satisfiable?

This is a decision (true/false) version of the corresponding approximation problem (find an assignment which comes close to maximizing the number of clauses satisfied). In a celebrated result of the early 1990s, it was shown that this problem is NP-hard; a good starting point for further details is a survey article by Arora and Lund (1996).

Suppose we are given a formula $\phi = C_1 \wedge C_2 \wedge \cdots \wedge C_m$ over variables $x_1, \ldots, x_n$, and with the stated guarantees. We will show that there is some constant $c > 1$ (which depends only upon $\epsilon$) such that any algorithm which can learn a 2-polytree within a multiplicative factor $c$ of optimal (in terms of log-likelihood) can also be used to decide whether $\phi$ is satisfiable.

For this reduction we construct a probability distribution out of the formula $\phi$. The distribution will be specified by the following probabilistic net.

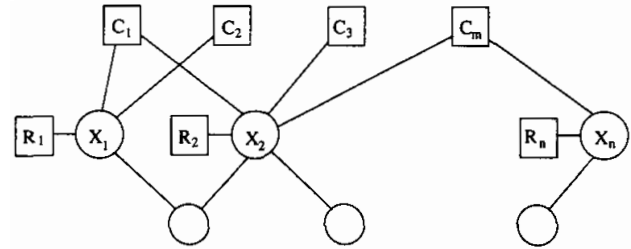

The edges in this graph represent correlations. Circles denote regular nodes. The squares are slightly more complicated structures whose nature will be disclosed later; for the time being they should be thought of as no different from the circles. There are three layers of nodes.

(1) The top layer consists of i.i.d. $B(p)$ random variables, one per clause; here "$B(p)$" denotes a Boolean random variable with probability $p$ of being one, that is, a coin with probability $p$ of coming up heads. This entire layer has entropy $mH(p)$.

(2) The middle layer contains two nodes per variable $x_i$: a principal node $X_i$ which is correlated with the nodes for the three $C_j$'s in which $x_i$ appears, and a $B(1/2)$ satellite node called $R_i$. Let us focus on a particular variable $x_i$ which appears in clauses $C_1, C_2$ with one polarity and $C_3$ with the opposite polarity – this is the general case since each variable appears exactly three times. For instance, $C_1$ and $C_2$ might contain $x_i$ while $C_3$ contains $\bar{x}_i$. Then the nodes $R_i, X_i$ will have the following structure.

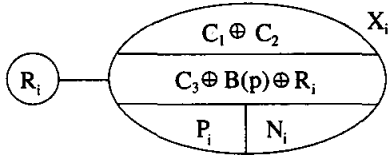

Here $P_i$ and $N_i$ are both $B(1/2)$, and stand for "previous" and "next." Note each $R_i$ is Boolean with entropy one and each $X_i$ is a four-tuple of Boolean values, with entropy $H(2p(1-p)) + 3$. Thus the second layer of the graph has overall entropy $n(H(2p(1-p)) + 4)$.

(3) The third layer (of $n - 1$ nodes) joins together consecutive nodes $X_i, X_{i+1}$ of the second layer. The value of the $i$th node in this level, $1 \leq i \leq n - 1$, is $N_i \oplus P_{i+1}$, and the entire level has entropy $n - 1$.

To sample randomly from this distribution, pick values $C_1, \ldots, C_m, R_1, \ldots, R_n, P_1, \ldots, P_n, N_1, \ldots, N_n$ independently and at random, and then set $X_1, \ldots, X_n$ based upon these.

Suppose that the learning algorithm has so much data that there is no sampling error. Thus it knows the exact entropies $H(\cdot)$ and conditional entropies $H(\cdot|\cdot)$ for all combinations of variables. What kind of polytree can it construct? We will show that:

- If $\phi$ is satisfiable then there is a 2-polytree whose cost is some value $H^*$ which can easily be computed from $\phi$.

- If $\phi$ is not satisfiable (and so at most $m(1 - \epsilon)$ clauses can be satisfied by any assignment) then there is no 2-polytree of cost less than $cH^*$.

Thus, any algorithm which can find a 2-polytree within factor $c$ of optimal, can also decide whether $\phi$ is satisfiable.

How can we prove this? A good polytree must somehow tell us a satisfying assignment for $\phi$. If the polytree that is learned has an edge from $C_j$ to $X_i$, we will say "$C_j$ chooses $X_i$" and will take this as meaning that clause $C_j$ is satisfied on account of variable $x_i$. There are several steps towards showing this is well-defined.

First we will force the square nodes to have no inedges. This is quite easy to achieve; for instance, $R_1$, ostensibly a single coin flip, can be rigged to avoid inedges by giving it two default inedges via the following little gadget.

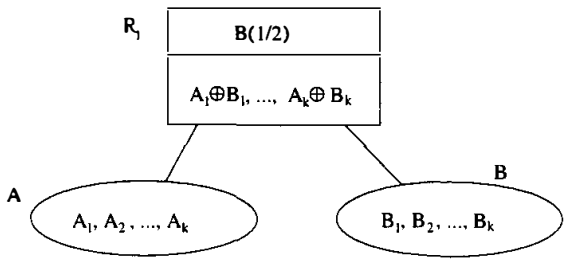

Here $A$ and $B$ are two extra nodes, independent of everything but $R_1$, which each contain $k$ i.i.d. copies of $B(q)$, called $A_1, A_2, \ldots, A_k$ and $B_1, \ldots, B_k$, for some constants $q < 1/2$ and $k > 1$. $R_1$ has now expanded to include $A_1 \oplus B_1, \ldots, A_k \oplus B_k$. On account of the polytree constraint, there can be at most two edges between $A$, $B$, and $R_1$. The optimal choice is to pick edges from $A$ and $B$ to $R_1$. Any other configuration will increase the cost by at least $k(H(2q(1-q)) - H(q)) > 1$ (the constant $k$ is chosen to satisfy this). These suboptimal configurations might permit inedges from outside nodes which reveal something about the $B(1/2)$ part of $R_1$; however such edges can provide at most one bit of information and will therefore never be chosen (they lose more than they gain). In short, such gadgets ensure that square nodes will have no inedges from the nodes in the overall graph shown above.

Next we will show that any half-decent polytree must possess all $2(n - 1)$ connections between the second and third levels. Any missing connection between consecutive $X_i, X_{i+1}$ causes the cost to increase by one and in exchange permits at most one extra edge among the $X_i$'s and $C_j$'s, on account of the polytree constraint. However we will see that none of these edges can decrease the cost by one.

Therefore, all the consecutive $X_i$'s are connected via the third layer, and each clause-node $C_j$ in the first level can choose at most one variable-node $X_i$ in the second level (otherwise there will be a cycle).

We now need to make sure that the clauses which choose a variable do not suggest different assignments to that variable. In our example above, variable $x_i$ appeared in $C_1, C_2$ with one polarity and $C_3$ with the opposite polarity. Thus, for instance, we must somehow discourage edges from both $C_1$ and $C_3$ to $X_i$, since these edges would have contradictory interpretations (one would mean that $x_i$ is true, the other would mean that $x_i$ is false). The structure of the node $X_i$ imposes a penalty on this combination.

Choose $p$ so that $H(p) = 1/2$, and define $\delta \stackrel{\text{def}}{=} 1 - H(2p(1-p)) > 0$. Assume that we start with a structure that has no edges in the first and second layers (apart from the hidden edges in the square nodes). The goal of the learning algorithm is to add edges from the $C_j$'s and $R_i$'s to the $X_i$'s which will bring the cost down as much as possible. Which edges is it likely to add? What are the possible parents of the $X_i$ depicted above?

(a) Just $R_i$: the entropy of $X_i$ is decreased by $\delta$.

(b) $R_i$, and one of $C_1, C_2, C_3$: the entropy of $X_i$ is decreased by $1/2$.

(b) $C_1, C_2$: the entropy decrease is $1 - \delta$.

(c) $C_1, C_3$ or $C_2, C_3$: the entropy decrease is $1/2 - \delta$.

Thus $C_1$ and $C_3$ will not both choose variable $X_i$, and more generally, the assignment embodied in the edges from clause-nodes to variable-nodes will be well-defined. Suppose $m'$ clauses are satisfiable (either $m' = m$ or $m' \leq m(1 - \epsilon)$). Then the edges into the $X_i$'s will decrease the cost of the second layer by $m'(1/2 - \delta) + n\delta$.

In this way, if all clauses are satisfiable, then the optimal structure has some cost $H^* = \theta(m)$, whereas if only $(1 - \epsilon)m$ clauses are satisfiable, then the optimal cost is $H^* + \theta(\epsilon m) \geq cH^*$ for some constant $c > 1$ which depends only upon $\epsilon$ and not upon $\phi$. Thus PT(2) is hard to approximate within some fixed constant factor. ∎

This proof relies heavily upon the degree constraint as a source of hardness. However it should be possible to produce similar effects using just the polytree constraint, and a little imagination.

## 4 Directions for further work

We have seen that the optimal branching, in which each node has at most one parent, is a good approximation to the optimal polytree, in terms of log-likelihood. Any algorithm with a better performance guarantee must be able to give nodes two or more parents when necessary. What are good heuristics for doing this? Can a linear programming approach yield something here?

Polytrees are currently the most natural class of directed graphs for which efficient inference algorithms are known. But in fact the junction tree algorithm works quickly for any directed graph whose moralized, triangulated, undirected version has very small cliques. Is there a larger class of directed graphs with simple characterization which permits fast inference and for which efficient, provably good learning algorithms can be constructed?

## Acknowledgements

The author is grateful to Nir Friedman for introducing him to this field, and thanks the referees for their useful comments.